\PassOptionsToPackage{prologue,dvipsnames}{xcolor}
\pdfoutput=1
\documentclass[11pt]{article}

\usepackage[preprint]{acl}

\usepackage[dvipsnames]{xcolor}
\usepackage{times}
\usepackage{latexsym}

\usepackage[T1]{fontenc}
\usepackage{microtype}

\usepackage{inconsolata}
\usepackage{ulem}
\usepackage{graphicx}
\usepackage{subfigure}

\author{Zhi Zheng \\
  National University of Singapore  \\
  \texttt{zhengzhi@comp.nus.edu.sg} \\\And
  Wee Sun Lee \\
  National University of Singapore \\
  \texttt{leews@comp.nus.edu.sg} \\}
\usepackage{multirow}
\usepackage[T1]{fontenc}% use 8-bit T1 fonts
\usepackage{hyperref}% hyperlinks
\usepackage{url}  % simple URL typesetting
\usepackage{booktabs}% professional-quality tables
\usepackage{amsfonts}% blackboard math symbols
\usepackage{nicefrac}% compact symbols for 1/2, etc.
\usepackage{microtype}% microtypography
\usepackage{amssymb}
\usepackage{graphicx}
\usepackage{subcaption}
\usepackage{amsmath}
\usepackage{graphicx}
\usepackage{listings}
\usepackage[most]{tcolorbox}

\newtcolorbox{dialogbox2}[1][]{
  arc=4mm,
  colback=lightgray!20,
  colframe=green!25!black,
  rounded corners,
  boxrule=1.5pt,
  fonttitle=\sffamily\bfseries,
  coltitle=white,
  toptitle=2mm,
  bottomtitle=2mm,
  title=#1, % 使用传递的参数作为标题
  frame style={dashed}, % 使用虚线边框
  breakable, % 允许换页
}
\newtcolorbox{dialogbox}[1][]{
  colback=lightgray!40, % 加重底色
  colframe=darkgray!80, % 不要边框
  boxrule=0pt, % 边框宽度为0
  fonttitle=\sffamily\bfseries,
  coltitle=white,
  toptitle=2mm,
  bottomtitle=2mm,
  title=#1, % 使用传递的参数作为标题
  frame style={solid}, % 使用实线边框（虽然边框宽度为0）
  breakable % 允许换页
}
\definecolor{deepcyan}{RGB}{85, 142, 213}
\definecolor{deepred}{RGB}{192, 80, 77}
\usepackage{listings}
\usepackage{algorithm,algorithmic}
\usepackage{geometry}
\usepackage{wrapfig,lipsum,booktabs}
\usepackage{blindtext}
\title{Reasoning-CV: Fine-tuning Powerful Reasoning LLMs for Knowledge-Assisted Claim Verification}

\begin{document}

\maketitle
\begin{abstract}
Claim verification is essential in combating misinformation, and large language models (LLMs) have recently emerged in this area as powerful tools for assessing the veracity of claims using external knowledge. Existing LLM-based methods for claim verification typically adopt a \textit{Decompose-Then-Verify} paradigm, which involves decomposing complex claims into several independent sub-claims and verifying each sub-claim separately. However, this paradigm often introduces errors during the claim decomposition process. To mitigate these errors, we propose to develop the \textit{Chain-of-Thought (CoT)-Verify} paradigm, which leverages LLM reasoning methods to generate CoT-verification paths for the original complex claim without requiring decompositions into sub-claims and separate verification stages. The \textit{CoT-Verify} paradigm allows us to propose a natural fine-tuning method called Reasoning-CV to enhance the verification capabilities in LLMs. Reasoning-CV includes a supervised fine-tuning (SFT) stage and a self-improvement direct preference optimization (DPO) stage. Utilizing only an 8B pre-trained LLM, Reasoning-CV demonstrates superior knowledge-assisted claim verification performances compared to existing \textit{Decompose-Then-Verify} methods, as well as powerful black-box LLMs such as \textit{GPT-4o}+CoT and o1-preview. Our code is available\footnote[1]{The code of Reasoning-CV is available at \url{https://github.com/zz1358m/Reasoning-CV}.}.
\end{abstract}

\section{Introduction}
Claim verification \cite{aly2021feverous,jiang2020hover} (also known as fact-checking \cite{eldifrawi2024automated}) is a crucial task that involves evaluating the veracity of a complex claim using provided or retrieved knowledge as evidence. This task plays a vital role in curbing the spread of misinformation \cite{tambuscio2015fact,min2023factscore} and is instrumental in identifying hallucinations \cite{huang2023survey,nie2024facttest} of generative models. With the recent advancements in Large Language Models (LLMs) \cite{touvron2023llama,achiam2023gpt,liu2024deepseek}, LLM-based approaches \cite{wang2023explainable,zhao2024pacar} have demonstrated exceptional performance in the knowledge-assisted claim verification task.
\begin{figure}[htbp]
\centering
\includegraphics[width=0.96\linewidth]{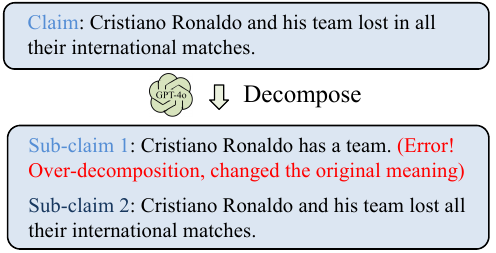}
\caption{An example of claim decomposition using \textit{GPT-4o} and prompts from \citet{min2023factscore}. In this case, we get a redundant sub-claim (i.e., \textcolor{deepcyan}{Sub-Claim 1} in Figure) that changes the original meaning.}
\label{fig:decomposition error}
\end{figure}

\begin{figure*}[htbp]
\centering
\subfigure[\textit{Decompose-Then-Verify} paradigm for knowledge-assisted claim verification]{\includegraphics[width=\textwidth]{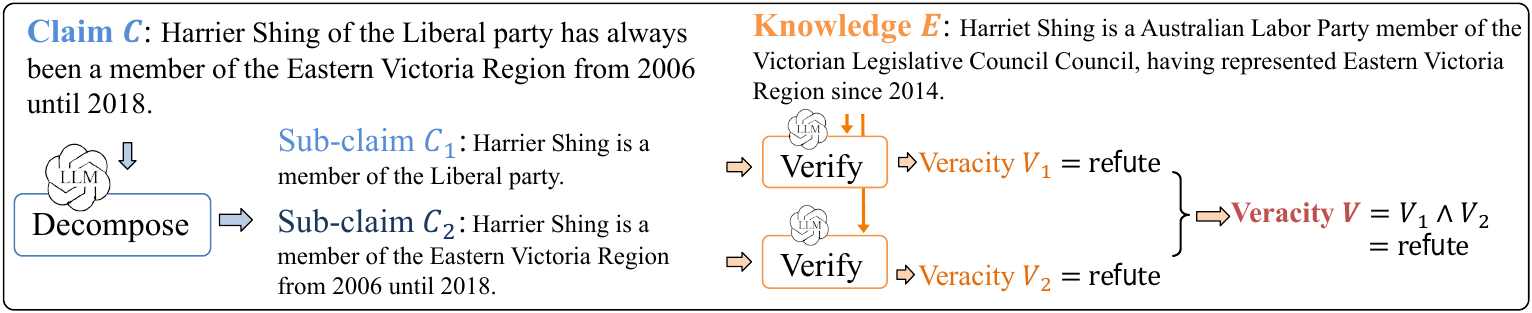}}
\subfigure[\textit{CoT-Verify} for end-to-end knowledge-assisted claim verification (Ours)]{\includegraphics[width=\textwidth]{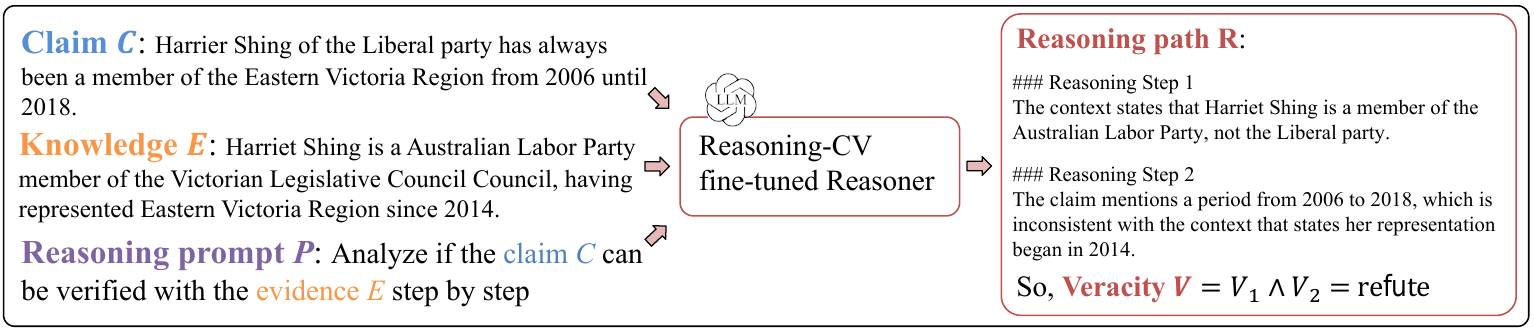}}\caption{Existing LLM-based claim verification methods generally adopt a \textit{Decompose-Then-Verify} paradigm (a). These methods first break a complex claim into several independent sub-claims, then leverage provided or retrieved knowledge to judge the veracity of each simple sub-claim. We propose to use a \textit{CoT-Verify} paradigm (b), which aims at directly generating high-quality CoT-verification paths for the veracity of complex claims. It can reduce the number of LLM calls, eliminate the decomposition error, and achieve significantly better accuracy after the proposed Reasoning-CV fine-tuning. }\label{fig:figure1}
\end{figure*}

As an intricate task, claim verification requires a precise understanding of complex claims and their associated knowledge. So, existing LLM-based methods generally adopt a \textit{Decompose-Then-Verify} paradigm \cite{hu2024decomposition} to reduce the complexity of claim verification. As shown in Figure \ref{fig:figure1}(a), these methods begin by leveraging the text-processing capabilities of LLMs to decompose a complex claim into several independent sub-claims. Then, LLMs are employed again to judge the veracity of each simple sub-claim, taking provided or retrieved knowledge as evidence \cite{wang2023explainable}. Finally, \textit{Decompose-Then-Verify} methods will judge the claim as correct if all its sub-claims are supported. Although this paradigm can simplify the verification of complex claim sentences, LLM-based decomposition will inevitably lead to errors, obtaining uncertain or wrong sub-claims  \cite{hu2024decomposition} (e.g., the case in Figure \ref{fig:decomposition error}), which will significantly impact the verification accuracy.

In this paper, we observe that when employing the Chain-of-Thought (CoT) prompt strategy \cite{wei2022chain,sprague2024cot} for claim verification, advanced LLMs such as \textit{GPT-4o} will naturally perform verification analysis on each of the key facts among the claim in its CoT reasoning path (See Appendix \ref{example} for examples). This observation indicates that LLMs have the potential to effectively analyze complex claims with a \textit{CoT-Verify} process in a single LLM call, and explicitly decomposing claims into separate verification stages is unnecessary. Compared to \textit{Decompose-Then-Verify}, the \textit{CoT-Verify} paradigm can reduce the number of LLM calls while mitigating the possible errors during the claim decomposition process.

To develop powerful reasoning LLMs for \textit{CoT-Verify}, this paper proposes a novel \textbf{Reasoning-CV} (Reasoning-Claim-Verification) fine-tuning method for open-source LLMs. As shown in Figure \ref{fig:pipeline1}, the proposed Reasoning-CV method is a \textbf{two-stage fine-tuning process}. In the first stage, we prompt \textit{GPT-4o} to generate reliable reasoning paths for ground-truth labels and do supervised fine-tuning (SFT) on pre-trained LLMs. In the second stage, we design a self-improvement direct preference optimization (DPO) \cite{rafailov2023direct} process, gradually guiding the fine-tuned LLM to update the consistency and judgment of its reasoning path. We evaluate the proposed Reasoning-CV on a wide collection of knowledge-assisted claim verification test sets and benchmarks, and LLMs fine-tuned with Reasoning-CV demonstrate superior performance compared to existing claim verification methods and frontier LLMs including \textit{GPT-4o} and o1-preview, utilizing only 8B or fewer parameters. Our contributions are summarized as follows:
\begin{itemize}
\item This paper first proposes to build the claim verification process in a long CoT reasoning path and presents Reasoning-CV to fine-tune LLMs for high-quality reasoning paths. 
\item The proposed Reasoning-CV presents a novel method for generating high-quality 
%initial SFT training data as well as 
DPO training data, which is used to fine-tune LLMs in an iterative self-improvement manner. Instead of simply generating CoT-verification paths and checking whether they agree with the ground truth verification label, we provide the possible labels and ask the LLM to generate a CoT that agrees with each label. Conditioning on the label to generate allows LLMs to generate higher-quality CoTs for correct labels and confusing CoTs (that we want to learn not to generate) for incorrect labels.
\item Reasoning-CV can achieve superior knowledge-assisted claim verification performance compared to existing methods and frontier LLMs, using only a \textit{Meta-LlaMA-3-8B-Instruct} base LLM.
\end{itemize}

\section{Related Work}
\subsection{Task Definition: Claim Verification} \label{definition}

The claim verification task aims to determine the veracity $V$ of a claim $C$ based on knowledge $E$. There are two settings for the source of knowledge $E$, gold evidence, and open book \cite{aly2021feverous}. In the \textbf{gold evidence} setting, each claim is provided with knowledge that can determine its veracity, while the \textbf{open book} setting requires verification methods to retrieve knowledge from sources based on the claim $C$. There are also two settings for the range of veracity $V$, that is, \textbf{\textit{w/o} NEI} and \textbf{\textit{w} NEI}. Under the \textbf{\textit{w/o} NEI} setting, the veracity is predicted from \textit{`support`} and \textit{`refute`}, while the \textbf{\textit{w} NEI} setting introduces another \textit{`not enough evidence`} (abbreviated as \textit{`NEI`}) option. Some datasets, like FEVEROUS \cite{aly2021feverous}, have no \textit{`NEI`} label and support only the \textbf{\textit{w/o} NEI} setting \cite{jafari2024robust}, while other datasets (i.e., Healthver \cite{Sarrouti2021Healthver}) can support the \textbf{\textit{w} NEI} setting.

\subsection{\textit{Decompose-Then-Verify} Paradigm}

The \textit{Decompose-Then-Verify} paradigm is widely adopted in LLM-based claim verification methods \cite{hu2024decomposition}, including FACTSCORE \cite{min2023factscore}, SAFE \cite{wei2024long}, PACAR \cite{zhao2024pacar}, etc. \cite{wang2023explainable, zhang2023towards,jafari2024robust}. Their verification processes involve breaking down the claim $C$ into a set of sub-claims $\mathcal{S} = \{c_1,c_2,\ldots,c_{p}\}$ with LLMs and using a verifier (usually an LLM-based verifier \cite{zhao2024pacar}) to assess the veracity $v_i$ of each sub-claim $c_i,\ i\in \{1,\ldots,p\}$. Finally, individual verification results are aggregated to produce a final judgment of veracity $V$. For example, in FOLK \cite{wang2023explainable}, the final veracity is processed as the conjunction paradigm for the veracity of sub-claims as follows:
\begin{equation}
V \leftarrow v_1\wedge v_2 \wedge \ldots \wedge v_p.\label{folk}
\end{equation}
Compared to directly verifying the claim with reasoning steps (i.e., \textit{CoT-Verify}), \textit{Decompose-Then-Verify} methods can avoid verifying complex claim sentences \cite{min2023factscore}. However, as shown in recent research \cite{hu2024decomposition,tang2024minicheck}, these methods introduce decomposition errors, which can result in generating uncertain or inconsistent sub-claims (as shown in Figure \ref{fig:decomposition error}). Once making an incorrect decomposition, even the optimal sub-claim verification results may achieve an incorrect veracity judgment $V$, which will significantly impact the claim verification performance.

\begin{figure*}[htbp]
\centering
\subfigure[Stage 1: SFT]{\includegraphics[width=0.3257\textwidth]{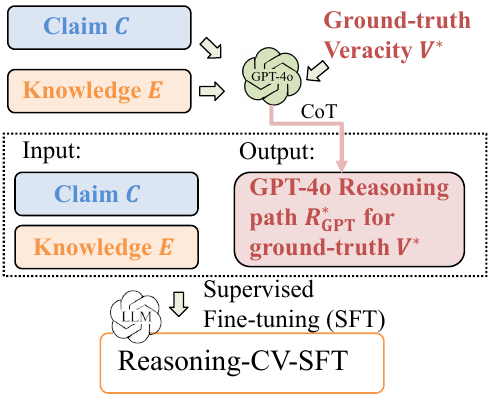}}
\qquad\qquad
\subfigure[Stage 2: Self-Improvement DPO]{\includegraphics[width=0.5\textwidth]{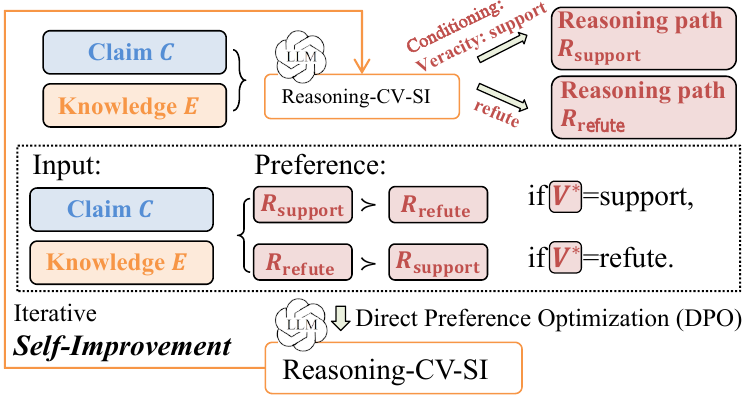}}\caption{The proposed Reasoning-CV is a two-stage fine-tuning method for knowledge-assisted claim verification. Given a training dataset with Claim $C$, Knowledge $E$, and ground-truth veracity $V^*$, Reasoning-CV can obtain LLMs with high-quality CoT-verification paths. The first stage aims at distilling the \textit{GPT-4o} generated reasoning path for ground-truth veracity $V^*$, and the second stage is designed to iteratively enhance the judgment and consistency of reasoning paths from the fine-tuned LLMs. }\label{fig:pipeline1}
\end{figure*}

\subsection{Other Paradigms}

Besides the \textit{Decompose-Then-Verify} paradigm, there are also claim verification methods with an adaptive retrieval and verification framework \cite{pan2023fact,shao2023enhancing,quelle2024perils}. Similar to adaptive retrieval methods for Retrieval-augmented generation (RAG) \cite{gao2023retrieval,asai2023self}, these methods first use the original claim $C$ to retrieve evidence for verification and then repeatedly handle the uncertain parts of the claim in the previous verification step by retrieving new knowledge as verification evidence. Compared to these methods, Reasoning-CV focuses on fine-tuning LLMs to make the first verification process sufficiently reliable. 

Directly employing the pre-trained LLMs with CoT prompts (e.g., \textit{GPT-4o}+CoT) or using black-box reasoning LLMs (e.g., o1-preview) are existing works under the \textit{CoT-Verify} paradigm. Some pre-trained reasoning models like o1-preview are proficient in handling complex claims $C$ \cite{sprague2024cot} without requiring decomposition or iterative processes, albeit at a higher cost. 

\begin{table*}[t]
\centering
\renewcommand\arraystretch{1.05}
\setlength{\tabcolsep}{2.5mm}
\resizebox{\textwidth}{!}{
\begin{tabular}{l|lll|c|c}
\bottomrule[0.5mm]
Dataset& Domain   & Claim& Evidence & \multicolumn{1}{c}{Veracity Label $V^*\in$}& \multicolumn{1}{|l}{Training data number} \\ \midrule[0.3mm]
FEVEROUS   & Wikipedia& Brief, Complex   & Tabular, Textual & \{\textit{support},\textit{refute}\}  & 26,828   \\
\multirow{3}{*}{HOVER} & \multirow{3}{*}{Wikipedia} & 2-hop claims& \multirow{3}{*}{Textual}&  & 9,006   \\
&& 3-hop claims&  & \{\textit{support}, \textit{refute}\} & 6,053  \\
&& 4-hop claims (complex) & &   & 3,012  \\
Healthver  & COVID-19& Brief, Complex & Textual& \{\textit{support}, \textit{NEI},  \textit{refute}\}& 10,590   \\ \hline
Total& All domains& Mixed Hardness   & Language &  \textit{w} NEI and  \textit{w/o} NEI & 55,489\\ \toprule[0.5mm]
\end{tabular}
}
\caption{The collection of the training dataset for Reasoning-CV. The Reasoning-CV training dataset covers three datasets with complex claims. Specifically, the Healthver dataset has a \textit{`NEI`} option.}
\label{data}
\end{table*}

\subsection{Fine-Tuning LLMs for Claim Verification}

Several claim verification methods propose fine-tuning techniques on neural models for counter-example generation \cite{zhu2023explain}, evidence retrieval \cite{zhang2024reinforcement,huang2024training}, or veracity generation between sub-claims and evidence \cite{zeng2024maple,tang2024minicheck}. Among these, Minicheck \cite{tang2024minicheck} is closely related to our work, as it fine-tunes LLMs mainly for the verification step in the \textit{Decompose-Then-Verify} framework, focusing on verifying simple claims based on provided evidence. In contrast, our proposed Reasoning-CV fine-tuning method enhances the reasoning abilities of LLMs, enabling them to tackle \textbf{complex} claims directly without decomposition, thereby avoiding errors explicitly associated with the decomposition process.

\section{Reasoning-CV: Obtaining Advanced Reasoning LLMs for \textit{CoT-Verify}}\label{method}

This paper proposes a two-stage fine-tuning method, Reasoning-CV, to develop powerful reasoning LLMs for knowledge-assisted claim verification. To optimize the quality of \textit{CoT-Verify} reasoning paths for both \textit{w} NEI and \textit{w/o} NEI settings, we first build a training dataset containing \textbf{55K} pairs of \textbf{claims} $C$ with different hardness and their \textbf{gold evidence knowledge} $E$ in text or tabular form. As shown in Table \ref{data}, the training data is collected from three well-known claim verification datasets with complex claims (that is, HOVER \cite{jiang2020hover}, FEVEROUS \cite{aly2021feverous}, and Healthver \cite{Sarrouti2021Healthver}). Among them, we include HOVER and FEVEROUS for the \textit{w/o} NEI setting and Healthver for the \textit{w} NEI setting.

As shown in Figure \ref{fig:pipeline1}, the proposed Reasoning-CV is a two-stage fine-tuning process. In the first stage, it performs SFT on an open-source LLM to distill the CoT-verification path from \textit{GPT-4o}. We provide \textit{GPT-4o} the ground-truth veracity (noted as $V^*$) for reliable CoT-verification paths. In the second stage, we ask the fine-tuned LLM to generate reasoning paths for each veracity option and use the generated reasoning paths to build preference pairs for DPO fine-tuning, emphasizing reasoning paths with the ground-truth veracity.

\subsection{First Stage: Reasoning Path SFT}

Compared to advanced black-box LLMs like \textit{GPT-4o}, pre-trained open-source LLMs like the 8B LlaMA LLM tend to generate inferior reasoning paths (Refer to the \textit{\textit{Meta-LlaMA-3-8B-Instruct} + CoT Prompt} results in Table \ref{result_nei}). So, the first stage of Reasoning-CV fine-tunes open-source LLMs with reliable reasoning paths generated by \textit{GPT-4o} to build their \textit{CoT-Verify} capability. 

To generate reliable \textit{GPT-4o} reasoning paths for claim $C$ and knowledge $E$, we require \textit{GPT-4o} to generate a CoT that agrees with the ground-truth label $V^*$. The generated reasoning path can be regarded as \textit{GPT-4o}'s CoT-explanation of the ground-truth label $V^*$. In our implementation, we prompt \textit{GPT-4o} with $P_{CE}$ for claim $C$ and knowledge $E$, and a conditioning prompt $P_{V^*}$ providing the ground truth label $V^*$. The \textit{GPT-4o} reasoning path $R^*_{GPT}$ is generated as follows (Refer to Appendix \ref{prompt} for detailed prompts and instructions):
\begin{equation}
R^*_{GPT} = \textit{GPT-4o}(P_{CE}, P_{V^*}).\label{stage1}
\end{equation}
We enumerate data in the training dataset (i.e., Table \ref{data}), forming a dataset of reasoning paths $R^*_{GPT}$. Although provided with the ground-truth veracity $V^*$, \textit{GPT-4o} will still generate reasoning paths with incorrect judgments. So, we conduct a data cleaning process that removes all these reasoning paths that do not match the label (i.e., $V^*\notin R^*_{GPT}$) and ultimately obtain 50,011 reliable paths. We use SFT to fine-tune LLMs over reliable \textit{GPT-4o} reasoning paths for three epoches and note the LLM after the first stage of Reasoning-CV as $LLM_{SFT}$.

\begin{figure}[t]
\centering
\includegraphics[width=\linewidth]{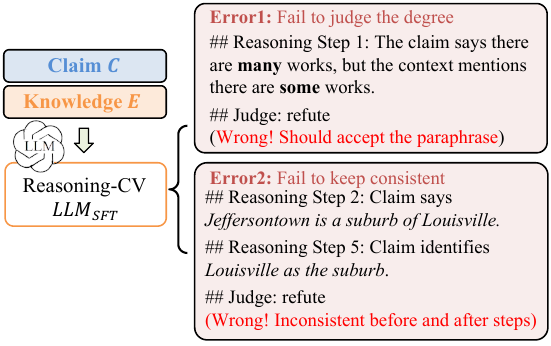}
\caption{A qualitative study for the possible error of reasoning LLMs after the first stage of Reasoning-CV fine-tuning. LLMs still face difficulties in judging the acceptance threshold (e.g., \textcolor{deepred}{Error1}) and in maintaining consistency for their long reasoning path (e.g., \textcolor{deepred}{Error2}). }
\label{fig:error}
\end{figure}

\subsection{Second Stage: Self-Improvement DPO}

As shown in Figure \ref{fig:error}, after the SFT stage, the fine-tuned $LLM_{SFT}$ will still have difficulties in: \textbf{1)} Judging the acceptance threshold. For example, in Case 1 of Appendix \ref{case_study}, the knowledge contains a fact using the adverb "some" to show its degree, and the claim paraphrases the word "some" to "many". Human judgment may accept this paraphrase, but $LLM_{SFT}$ incorrectly judges it as \textit{`refute`}. \textbf{2)} Keeping consistent in the former and latter parts of the reasoning path. When processing complex claims, for example, in Case 2 of Appendix \ref{case_study}, the reasoning path may initially provide correct assertions (Step 2). However, in step 5, $LLM_{SFT}$ incorrectly changes its previous assertions, resulting in inconsistency and a wrong judgment. 

We believe both drawbacks stem from the fact that the generated \textit{GPT-4o} reasoning paths $R^*_{GPT}$ are not optimal for the \textit{CoT-Verify} of claims. So, to further improve the judgment and consistency of fine-tuned LLMs, we propose the second stage of Reasoning-CV, iteratively building preferences for correct over wrong reasoning paths.

In the second stage of Reasoning-CV, we provide the fine-tuned LLM with all possible veracity options (i.e., \textit{`support`} and \textit{`refute`} for data in FEVEROUS and HOVER; \textit{`support`}, \textit{`NEI`}, and \textit{`refute`} for data in Healthver) and ask the LLM to conditionally generate reasoning paths for each option. Then, we utilize the generated paths to fine-tune LLMs with the DPO technique \cite{rafailov2023direct}, preferring the reasoning path with the ground-truth veracity over incorrect ones. By employing LLMs after the DPO fine-tuning to generate new preference pairs, this conditional-generation-based DPO process can be conducted in a self-improving manner. So, starting from $LLM_{SI_0}=LLM_{SFT}$, in the round $i \in \{1,2,...\}$, we prompt the fine-tuned LLM $LLM_{SI_{i-1}}$ with $P_{CE}$ and the conditioning prompt (e.g., $P_{support}$, $P_{refute}$) to generate the preference dataset $\mathcal{D}_{SI_{i-1}}$ for DPO as follows (taking claims for FEVEROUS and HOVER as examples, which have only two veracity options, \textit{`support`} and \textit{`refute`}):
\begin{equation}
\begin{aligned}
&R_{support} = LLM_{SI_{i-1}}(P_{CE}, P_{support}),\\ &R_{refute} = LLM_{SI_{i-1}}(P_{CE}, P_{refute}).
\end{aligned}\label{Stage2}
\end{equation}
\begin{equation}
\begin{aligned}
&\mathcal{D}_{SI_{i-1}} = \\
&\quad \Big\{[V^* \in R_{support}](R_{refute}\prec R_{support} ),\\ &\qquad [V^* \in R_{refute}](R_{support}\prec R_{refute} )\Big\}.
\end{aligned}
\end{equation}
Similar to the first stage, we use the 55K Reasoning-CV training dataset for claims, evidence, and veracity labels in each round. For the Healthver dataset, we enumerate the \textit{`support`}, \textit{`NEI`}, and \textit{`refute`} options and Appendix \ref{nei_SI} shows the details of constructing the preference $\mathcal{D}_{SI_{i-1}}$ for Healthver. The fine-tuned LLMs' explanation of incorrect veracities tends to be forced and inconsistent, and using DPO to discourage such reasoning paths can help LLMs learn the thresholds for judgments and help improve the consistency of the \textit{CoT-Verify} generation. We conduct two rounds of self-improvement by default ($i\in\{1,2\}$), and in each round of obtaining $LLM_{SI_i}$, we fine-tune $LLM_{SI_{i-1}}$ on the preference dataset of paths $\mathcal{D}_{SI_{i-1}}$ for three epoches.

\begin{table*}[t]
\centering
\renewcommand\arraystretch{1.0}
\setlength{\tabcolsep}{1.25mm}
\resizebox{\textwidth}{!}{
\begin{tabular}{lcccccccc}
\bottomrule[0.5mm]
\multicolumn{9}{c}{Gold Evidence} \\ \hline
\multicolumn{1}{c|}{Dataset}& \multicolumn{1}{c|}{Param Size} & \multicolumn{1}{c|}{FEVEROUS}& HOVER (2-hop)& HOVER (3-hop)& \multicolumn{1}{c|}{HOVER (4-hop)}& \multicolumn{2}{c|}{Healthver}& \multicolumn{1}{l}{} \\ \cline{1-8}
\multicolumn{1}{c|}{Method}& \multicolumn{1}{c|}{Settings}   & \multicolumn{1}{c|}{\textit{w/o NEI}} & \textit{w/o NEI} & \textit{w/o NEI} & \multicolumn{1}{c|}{\textit{w/o NEI}} & \textit{w/o NEI} & \multicolumn{1}{c|}{\textit{w NEI}} & \textit{Avg.}  \\ \hline
\multicolumn{1}{l|}{Meta-LLaMA-3-70B-Instruct} & \multicolumn{1}{c|}{70B}   & \multicolumn{1}{c|}{91.17}  & 76.28  & 73.15  & \multicolumn{1}{c|}{69.01}  & 71.44  & \multicolumn{1}{c|}{\underline{64.44}}& 74.25 \\
\multicolumn{1}{l|}{GPT-4o}& \multicolumn{1}{c|}{N/A}   & \multicolumn{1}{c|}{91.64}  & 79.48  & 75.83  & \multicolumn{1}{c|}{73.52}  & 67.87  & \multicolumn{1}{c|}{58.37}& 74.45 \\
\multicolumn{1}{l|}{GPT-4o + CoT Prompt}  & \multicolumn{1}{c|}{N/A}   & \multicolumn{1}{c|}{90.80}  & 81.44  & 76.32  & \multicolumn{1}{c|}{74.86}  & 70.26  & \multicolumn{1}{c|}{52.64}& 74.39 \\
\multicolumn{1}{l|}{o1-preview} & \multicolumn{1}{c|}{N/A}   & \multicolumn{1}{c|}{89.43}  & \underline{83.27} & \underline{78.69} & \multicolumn{1}{c|}{\underline{79.59}} & 68.84  & \multicolumn{1}{c|}{62.32}& \underline{77.02}\\
\multicolumn{1}{l|}{Minicheck-7B}& \multicolumn{1}{c|}{7B}& \multicolumn{1}{c|}{89.54}  & 82.62  & 71.68  & \multicolumn{1}{c|}{67.04}  & 55.63  & \multicolumn{1}{c|}{-}& 73.30 \\ \hline
\multicolumn{9}{c}{Estinblished \textit{Decompose-Then-Verify} Methods}  \\ \hline
\multicolumn{1}{l|}{PACAR}& \multicolumn{1}{c|}{N/A}   & \multicolumn{1}{c|}{\underline{94.43*}} & 76.86*  & 70.10*  & \multicolumn{1}{c|}{69.95*}  & - & \multicolumn{1}{c|}{-}& -\\
\multicolumn{1}{l|}{FactScore}  & \multicolumn{1}{c|}{N/A}   & \multicolumn{1}{c|}{85.55}  & 79.00  & 68.12  & \multicolumn{1}{c|}{62.49}  & 56.31  & \multicolumn{1}{c|}{60.93}& 68.73 \\
\multicolumn{1}{l|}{Decompose + Minicheck-1B}  & \multicolumn{1}{c|}{1B}& \multicolumn{1}{c|}{84.80}  & 75.47  & 61.67  & \multicolumn{1}{c|}{58.81}  & 47.35  & \multicolumn{1}{c|}{-}& 65.62 \\
\multicolumn{1}{l|}{Decompose + Minicheck-7B}  & \multicolumn{1}{c|}{7B}& \multicolumn{1}{c|}{85.66}  & 78.39  & 61.67  & \multicolumn{1}{c|}{53.11}  & 49.53  & \multicolumn{1}{c|}{-}& 65.67 \\ \hline
\multicolumn{9}{c}{Reasoning-CV for High-Qualty \textit{CoT-Verify} }\\ \hline
\multicolumn{1}{l|}{Meta-LLaMA-3-8B-Instruct}  & \multicolumn{1}{c|}{8B}& \multicolumn{1}{c|}{90.34}  & 74.24  & 67.21  & \multicolumn{1}{c|}{66.39}  & \underline{71.72} & \multicolumn{1}{c|}{55.48}& 70.90 \\
\multicolumn{1}{l|}{+ CoT Prompt}& \multicolumn{1}{c|}{8B}& \multicolumn{1}{c|}{87.06}  & 74.10  & 67.29  & \multicolumn{1}{c|}{64.09}  & 62.19  & \multicolumn{1}{c|}{52.06}& 67.80 \\
\multicolumn{1}{l|}{+ Reasoning-CV-$LLM_{SFT}$}   & \multicolumn{1}{c|}{8B}& \multicolumn{1}{c|}{93.72}  & 81.26  & 78.70  & \multicolumn{1}{c|}{75.65}  & 73.10  & \multicolumn{1}{c|}{64.19}& 77.77 \\
\multicolumn{1}{l|}{+ Reasoning-CV-$LLM_{SI_1}$} & \multicolumn{1}{c|}{8B}& \multicolumn{1}{c|}{95.22}  & 84.01  & 83.06  & \multicolumn{1}{c|}{81.01}  & 76.40  & \multicolumn{1}{c|}{\textbf{67.72}} & 81.24 \\
\multicolumn{1}{l|}{+ Reasoning-CV-$LLM_{SI_2}$} & \multicolumn{1}{c|}{8B}& \multicolumn{1}{c|}{\textbf{95.50}}   & \textbf{85.97}   & \textbf{83.93}   & \multicolumn{1}{c|}{\textbf{83.02}}   & \textbf{78.41}   & \multicolumn{1}{c|}{65.59}& \textbf{82.07}  \\ \toprule[0.5mm]
\end{tabular}
}
\caption{Results of Reasoning-CV fine-tuned LLMs in in-domain test sets with gold evidence considering both \textit{w NEI} and \textit{w/o NEI} settings. \textit{Avg.} shows the average performance of methods on different test sets and settings. We \textbf{bold} the best-performing LLM on each test set and \underline{underlines} the best-performing baseline. Results with * are collected from \citet{zhao2024pacar}.}
\label{result_nei}
\end{table*}

\section{Experiment}

In this section, we apply the proposed Reasoning-CV fine-tuning method to open-source LLM \textit{Meta-LlaMA3.1-8B-Instruct} and evaluate the ability of Reasoning-CV to obtain reasoning LLMs for powerful knowledge-assisted \textit{CoT-Verify}.

\paragraph{Dataset}
In this section, we include three in-domain test sets from datasets with complex claims discussed in Table \ref{data}, i.e., \textbf{FEVEROUS} \cite{aly2021feverous}, \textbf{HOVER} \cite{jiang2020hover}, and \textbf{Healthver} \cite{Sarrouti2021Healthver}. Meanwhile, to evaluate the generalization ability of the fine-tuned LLMs on claim-evidence pairs of different domains, we also include \textbf{LLM-AggreFact} \cite{tang2024minicheck}, \textbf{SciFact} \cite{wadden2020fact}, and \textbf{VitaminC} \cite{schuster2021get} datasets for out-of-domain. These datasets consist of relatively simple claims collected from a wide range of fields, including LLM generations \cite{tang2024minicheck} and science \cite{wadden2020fact}, which are rarely covered in the Reasoning-CV training dataset. Appendix \ref{dataset_info} includes details for these test sets.
% Moreover, to test the generalization of the fine-tuned LLM to different knowledge sources, we consider the gold evidence setting as well as the open book setting.

\paragraph{Implementation Details}

As mentioned in Section \ref{definition}, FEVEROUS, HOVER, and LLM-AggreFact only support the \textit{w/o} NEI setting. Datasets Healthver, VitaminC, and SciFact have the \textit{`NEI`} option in their label, so they can adopt both the \textit{w} NEI setting and the \textit{w/o} NEI setting (by considering the \textit{`NEI`} label as a special case of \textit{`refute`}). For claim verification under the two settings, we utilize the same reasoning LLM with different prompts shown in Appendix \ref{prompt}. We implement both stages of Reasoning-CV fine-tuning on an H100-96GB GPU with LoRA \cite{hu2022lora}. The LoRA rank is 64 in our experiment, and the learning rates for the two stages are $5e^{-5}$ and $5e^{-6}$, respectively. 

\paragraph{Baseline and Metrics}

This section involves a wide range of knowledge-assisted claim verification methods as the baselines, which include \textbf{1)} Advanced LLMs and black-box reasoning models, i.e., \textbf{Minicheck} \cite{tang2024minicheck}, \textit{\textbf{GPT-4o}}, \textbf{\textit{GPT-4o}+CoT}, and \textbf{\textit{o1-preview}}. \textbf{2)} Claim verification methods with \textit{Decompose-Then-Verify} paradigm, e.g., \textbf{FactScore} \cite{min2023factscore}, \textbf{PACAR} \cite{zhao2024pacar}, and \textbf{Decompose + Minicheck \cite{tang2024minicheck}}. For Decompose + Minicheck, we first decompose complex claims using prompts in \citet{min2023factscore}, then introduce Minicheck \cite{tang2024minicheck} fine-tuned LLMs to verify each of them. Please refer to Appendix \ref{baseline_info} for the implementation details of baselines. 

Following \citet{zhao2024pacar}, this paper uses the Macro-F1 score as the evaluation metric to assess the methods' performance on test sets. To better exhibit the performance difference between methods, we multiply each Macro-F1 score by 100.

\begin{table*}[htbp]
\centering
\renewcommand\arraystretch{1.0}
\setlength{\tabcolsep}{1mm}
\resizebox{1\textwidth}{!}{
\begin{tabular}{lccccccccccc}
\bottomrule[0.5mm]
\multicolumn{12}{c}{Gold Evidence}   \\ \hline
\multicolumn{1}{c|}{Dataset}  & \multicolumn{1}{c|}{Param Size} & \multicolumn{1}{c|}{LLM-AggreFact}& \multicolumn{2}{c}{SciFact-train} & \multicolumn{2}{c|}{SciFact-dev}   & \multicolumn{2}{c}{VitaminC-dev}  & \multicolumn{2}{c|}{VitaminC-test} & \multicolumn{1}{l}{} \\ \cline{1-11}
\multicolumn{1}{c|}{Method}   & \multicolumn{1}{c|}{Settings}   & \multicolumn{1}{c|}{\textit{w/o NEI}} & \textit{w/o NEI} & \textit{w NEI} & \textit{w/o NEI} & \multicolumn{1}{c|}{\textit{w NEI}} & \textit{w/o NEI} & \textit{w NEI} & \textit{w/o NEI} & \multicolumn{1}{c|}{\textit{w NEI}} & \textit{Avg.} \\ \hline
\multicolumn{1}{l|}{GPT-4o}   & \multicolumn{1}{c|}{N/A}& \multicolumn{1}{c|}{\underline{\textbf{79.40}}} & \underline{\textbf{88.99}} & \underline{77.57}& 85.57& \multicolumn{1}{c|}{\underline{75.50}}& \underline{81.48}  & \underline{65.21}& \underline{83.37}  & \multicolumn{1}{c|}{\underline{67.02}}& \underline{78.23}  \\
\multicolumn{1}{l|}{o1-preview}   & \multicolumn{1}{c|}{N/A}& \multicolumn{1}{c|}{-}& 86.53& 81.44  & 86.44& \multicolumn{1}{c|}{79.72}  & -& -  & -& \multicolumn{1}{c|}{-}  & -\\
\multicolumn{1}{l|}{Minicheck-1B} & \multicolumn{1}{c|}{1B} & \multicolumn{1}{c|}{67.86}& 81.87& -  & 82.57& \multicolumn{1}{c|}{-}  & 71.66& -  & 73.77& \multicolumn{1}{c|}{-}  & -\\
\multicolumn{1}{l|}{Minicheck-7B} & \multicolumn{1}{c|}{7B} & \multicolumn{1}{c|}{76.58}& 86.71& -  & \underline{87.21}  & \multicolumn{1}{c|}{-}  & 78.52& -  & 78.87& \multicolumn{1}{c|}{-}  & -\\ \hline
\multicolumn{12}{c}{Reasoning-CV for High-Qualty \textit{CoT-Verify} } \\ \hline
\multicolumn{1}{l|}{Meta-LLaMA-3-8B-Instruct} & \multicolumn{1}{c|}{8B} & \multicolumn{1}{c|}{74.58}& 86.34& 68.89  & 83.14& \multicolumn{1}{c|}{67.08}  & 75.50& 53.37  & 77.40& \multicolumn{1}{c|}{55.37}  & 71.30\\
\multicolumn{1}{l|}{+ CoT Prompt} & \multicolumn{1}{c|}{8B} & \multicolumn{1}{c|}{73.61}& 81.31& 66.56  & 83.81& \multicolumn{1}{c|}{62.19}  & 78.05& 62.10  & 78.93& \multicolumn{1}{c|}{63.31}  & 72.21 \\
\multicolumn{1}{l|}{+ Reasoning-CV-$LLM_{SFT}$}   & \multicolumn{1}{c|}{8B} & \multicolumn{1}{c|}{76.21}& 86.19& 78.99  & \textbf{87.46}& \multicolumn{1}{c|}{80.75}  & 82.38& 68.22  & 83.91& \multicolumn{1}{c|}{68.88}  & 79.22\\
\multicolumn{1}{l|}{+ Reasoning-CV-$LLM_{SI_1}$} & \multicolumn{1}{c|}{8B} & \multicolumn{1}{c|}{76.96}   & 87.82& \textbf{83.09} & 86.36   & \multicolumn{1}{c|}{\textbf{80.97}} & \textbf{83.18}   & \textbf{68.46} & \textbf{84.84}   & \multicolumn{1}{c|}{\textbf{69.44}} & \textbf{80.12}   \\
\multicolumn{1}{l|}{+ Reasoning-CV-$LLM_{SI_2}$} & \multicolumn{1}{c|}{8B} & \multicolumn{1}{c|}{77.10}   & 88.70& 82.99 & 86.44& \multicolumn{1}{c|}{80.16}  & 83.01& 66.29  & 84.81& \multicolumn{1}{c|}{67.35}  & 79.65\\ \toprule[0.5mm]
\end{tabular}}
\caption{Results of Reasoning-CV fine-tuned LLMs in out-of-domain test sets with gold evidence considering both \textit{w NEI} and \textit{w/o NEI} settings. We \textbf{bold} the best-performing LLM on each test set and \underline{underlines} the best-performing baseline. Results with * are collected from \citet{tang2024minicheck}.}\label{OOD}
\end{table*}

\subsection{In-Domain Performance}

We first investigate the effectiveness of the proposed Reasoning-CV on in-domain test sets (i.e., the accompanying test sets for FEVEROUS, HOVER, and Healthver shown in Table \ref{data}) with gold evidence. For Healthver, we evaluate Reasoning-CV with both \textit{w} NEI and \textit{w/o} NEI settings. As shown in Table \ref{result_nei}, when fine-tuning an 8B LLaMA base LLM, Reasoning-CV can significantly improve the claim verification performance stage-by-stage on all test sets with both \textit{w} NEI and \textit{w/o} NEI settings. Moreover, \textit{Meta-LLaMA-3.1-8B-Instruct} + Reasoning-CV-$LLM_{SI_2}$ can demonstrate superior average performances compared to advanced LLMs and \textit{Decompose-Then-Verify} baselines, including PACAR, Decompose + Minicheck-7B, \textit{GPT-4o} + CoT, even o1-preview. Results also indicate that the adopted \textit{CoT-Verify} paradigm can be better compared to the \textit{Decompose-Then-Verify} paradigm in knowledge-assisted claim verification.

\subsection{Generalize to Out-of-Domain Datasets}
The generalization ability of claim verification methods is significant, so we also evaluate Reasoning-CV fine-tuned LLMs on claims and knowledge collected from domains never seen in their fine-tuning. We consider five out-of-domain benchmarks and datasets, i.e., the LLM-AggreFact benchmark, the training set and the development set of SciFact, the development set and the test set of VitaminC. As shown in Table \ref{OOD}, utilizing an 8B base LLM, Reasoning-CV variants can achieve better results compared to Minicheck-7B and \textit{GPT-4o} on all five test sets, considering both \textit{w} NEI and \textit{w/o} NEI settings, demonstrating its generalization ability to a wide range of application scenarios.

%\subsection{Performance on Open Book Settings}

%In the Open Book setting, claim verification methods are required to retrieve knowledge $E$ from the Internet. To achieve this, we follow the idea from \citet{zhao2024pacar}, break the claim into sub-claims and retrieve the most related paragraph from Google with the Serper API, taking each sub-claim as the query. For the open book setting, we implement an 8B open book claim verification system in Appendix \ref{open_book_result}. The performances of Reasoning-CV and baselines with the open book setting are shown in Table \ref{result_nei_open}, where Reasoning-CV fine-tuned 8B LLMs can also outperform baselines, including \textit{GPT-4o}. However, we want to claim that, similar to \citet{tang2024minicheck}, Reasoning-CV focuses on improving the claim verification ability of LLMs under given knowledge. Generally, the evidence collected under the open book setting may not be enough to determine the veracity of claims \cite{vladika2024comparing}, so the open book setting cannot judge the superiority between methods.

\section{Discussion}

In this section, we conduct ablation studies for the settings of the proposed Reasoning-CV. We also discuss the open book setting and insights from our experiments.

\subsection{Ablation Studies}

To showcase the superiority of the Reasoning-CV settings and components, we conduct a series of ablation studies on the settings of each stage.

\paragraph{Ablation on the First-Stage Prompt $P_{V^*}$}

The key design in the first stage of Reasoning-CV is to provide \textit{GPT-4o} with the ground-truth veracity $V^*$, which may guide \textit{GPT-4o} to effectively understand the claim and generate better reasoning paths. To determine its significance, we design an ablation variant (\textit{w/o} Ground-truth $V^*$ in Table below) in which we remove the prompt $P_{V^*}$ in Equation (\ref{stage1}), sample \textit{GPT-4o} for up to 5 times with only $P_{CE}$ (a similar idea is used in \citet{wang2025rare}), and include the first sample with the correct veracity. As shown in Table \ref{ab1}, removing $P_{V^*}$ results in a significant decrease in model performance, especially in relatively hard datasets (e.g., HOVER and Healthver), so the original setting is a better choice.

\begin{table}[H]
\centering
\renewcommand\arraystretch{1}
\setlength{\tabcolsep}{1mm}
\resizebox{0.47\textwidth}{!}{
\begin{tabular}{l|ccc}
\hline
Method  & \multicolumn{1}{l}{FEVEROUS} & HOVER & Healthver \\ \hline
Original& 93.69& 78.50 & 74.11\\
\textit{w/o} Ground-truth $V^*$ & 92.91& 77.76 & 70.78\\ \hline
\end{tabular}}
\caption{Ablation on providing the ground-truth veracity in the first stage of Reasoning-CV. We consider only the \textit{w/o} NEI setting and take the average performance with 2 to 4 hops for HOVER.}\label{ab1}
\end{table}

\paragraph{Ablation on the Second-Stage Conditioning}
In the second stage of Reasoning-CV, we guide the LLMs $LLM_{SI_i}$ to generate with different veracity $V$, which can amplify the key difference between the chosen and rejected reasoning paths. To determine the effectiveness of generating conditioned on different veracities, we design an ablation variant (\textit{w/o} Conditioning in Table below) in which we remove the prompt $P_{support}$ and $P_{refute}$ in Equation (\ref{Stage2}) and sample $LLM_{SI_i}$ for two times with only $P_{CE}$. As shown in Table \ref{ab2}, removing the conditioning causes worse results. 
%The original setting guides the model to generate reasoning from 2 or 3 hypotheses (similar to the hypothesis-testing idea \cite{nie2024facttest}). This setting may promote the LLM's reasoning from the perspective of the alternation hypothesis, making the preferences of reasoning paths powerful.

\begin{table}[H]
\centering
\renewcommand\arraystretch{1}
\setlength{\tabcolsep}{2mm}
\resizebox{0.47\textwidth}{!}{
\begin{tabular}{l|ccc}
\hline
Method& \multicolumn{1}{l}{FEVEROUS} & HOVER & Healthver \\ \hline
Original  & 95.50& 84.31 & 78.41 \\
\textit{w/o} Conditioning & 94.44& 80.04 & 75.89\\ \hline
\end{tabular}}
\caption{Ablation on conditional generation with different veracities in the second stage. We consider only the \textit{w/o} NEI setting and take the average performance with 2 to 4 hops for HOVER.}\label{ab2}
\end{table}

\paragraph{Ablation on Pre-Trained LLMs}
We also apply Reasoning-CV on the \textit{LlaMA-3.2-3B-Instruct} LLM to verify its applicability on pre-trained LLMs of different sizes. The results of Reasoning-CV fine-tuned LLMs are shown in Table \ref{ab3}, where Reasoning-CV can significantly improve the performance in both in-domain (i.e., FEVEROUS) and out-of-domain (i.e., LLM-AggreFact) test sets stage-by-stage. The fine-tuned 3B LlaMA model can achieve competitive performance compared to \textit{GPT-4o} and Minicheck-7B on both test sets.

\begin{table}[H]
\centering
\renewcommand\arraystretch{1}
\setlength{\tabcolsep}{2mm}
\resizebox{0.485\textwidth}{!}{
\begin{tabular}{l|cc}
\hline
Method& \multicolumn{1}{l}{FEVEROUS} & \multicolumn{1}{l}{LLM-AggreFact} \\ \hline
GPT-4o& 91.64& 79.40\\
Minicheck-7B   & 89.54& 76.58\\
% 8B + Reasoning-CV-$LLM_{SI_2}$ & 95.73& 77.09\\ 
\hline
LlaMA-3B  & 86.69& 73.62\\
+ Reasoning-CV-$LLM_{SFT}$   & 92.59& 74.35\\
+ Reasoning-CV-$LLM_{SI_1}$ & 94.65& 75.98\\
+ Reasoning-CV-$LLM_{SI_2}$ & 94.75& 76.20\\ \hline
\end{tabular}}
\caption{Ablation on using 3B pre-trained LLM \textit{LlaMA-3.2-3B-Instruct} for Reasoning-CV. }\label{ab3}
\end{table}

\subsection{Performance on Open Book Settings}

As mentioned in Section \ref{definition}, besides the gold evidence setting used in Table \ref{result_nei} and Table \ref{OOD}, there is an open book setting in claim verification, which requires verifiers to retrieve knowledge $E$ from sources (e.g., the Internet). Usually, the collected evidence may not be enough to determine the veracity of claims \cite{vladika2024comparing}, so the veracity label for gold evidence may no longer be correct and cannot be used to reliably judge which claim verification method is better. Nonetheless, we evaluate the generalization ability of Reasoning-CV fine-tuned LLMs to the knowledge $E$ from this setting and provide the results in Appendix \ref{open_book_result}, where Reasoning-CV fine-tuned 8B LLMs can also outperform powerful baselines, including \textit{GPT-4o}.
%To achieve this, we follow the idea from \citet{zhao2024pacar}, break the claim into sub-claims and retrieve the most related paragraph from Google with the Serper API, taking each sub-claim as the query. For the open book setting, we implement an 8B open book claim verification system in Appendix \ref{open_book_result}. The performances of Reasoning-CV and baselines with the open book setting are shown in Table \ref{result_nei_open}, where Reasoning-CV fine-tuned 8B LLMs can also outperform baselines, including \textit{GPT-4o}. However, we want to claim that, similar to \citet{tang2024minicheck}, Reasoning-CV focuses on improving the claim verification ability of LLMs under given knowledge. 

\subsection{Comparison of the Second Stage to RL-based Fine-Tuning}

The idea of the proposed self-improvement DPO in Reasoning-CV is quite similar to reinforcement learning-based fine-tuning \cite{shao2024deepseekmath, liu2025understanding}. However, the inputted claim and knowledge for fine-tuning tends to be quite long (with usually thousands of tokens). Our efforts in fine-tuning LLMs for claim verification with RL ran into Out-of-Memory problems, failing to run on a single H100-96 GPU (same device for the current Reasoning-CV). Furthermore, Reasoning-CV is able to condition the generated reasoning sequences for DPO on the different target veracities, something not done in the usual RL algorithms.

%So, the self-improving DPO in Reasoning-CV can be a better choice to implement the idea of RL fine-tuning. It is significantly easier to implement and consumes fewer computational resources.

\subsection{Relation to Fine-Tuning Methods for Math Reasoning}\label{compare}

Currently, a large number of task-specific reasoning models have achieved outstanding results through distilling reasoning paths and fine-tuning LlaMA LLMs \cite{tu2025enhancing,li2025system,liu2025guardreasoner}. The main novelty of the proposed Reasoning-CV lies in its second stage. In the second stage, the conditional LLM generation for preferences improves the quality of the reasoning paths and calibrates the LLM's thresholds of judgments.

We believe the effectiveness of this design comes from the properties of the knowledge-assisted claim verification task, which has only two or three options for its judgment. Instead, for example, mathematical tasks tend to be open-ended with infinite possible answers, so we cannot implement the stage 2 of Reasoning-CV for these tasks. We believe that the stage 2 design in Reasoning-CV can be extended to any task with an enumerable number of answers, and we will consider it as future work.

\section{Conclusion}
In this paper, we propose to solve the knowledge-assisted claim verification task with the \textit{CoT-Verify} paradigm, building the claim verification process in a long CoT reasoning path. We present Reasoning-CV to fine-tune LLMs for high-quality reasoning paths. Reasoning-CV designs to generate reliable \textit{GPT-4o} reasoning paths in the first stage and representative preference datasets in the second stage, with conditioning on the label. It can significantly improve the 8B LLM's reasoning ability and get significantly better results on in-domain and out-of-domain test sets and benchmarks.

\section{Limitation}

As the main limitation of this paper, the reasoning ability in LLMs fine-tuned by Reasoning-CV cannot provide any help to the knowledge retrieval under the open book setting. In the future, we will try to integrate the retrieval process into the reasoning process. Moreover, as mentioned in Section \ref{compare}, we will also try to extend Reasoning-CV to a wider range of LLM-based tasks with enumerable answer options.

\bibliography{custom}

\appendix
\newpage
\section{Prompts for Claim Verification}\label{prompt}

In this section, we display all the prompts adopted for Reasoning-CV fine-tuning and the fine-tuned LLM. We use the Veracity Generation Prompt $P_{CE}$ in generating the veracity $V$ for claim $C$ and knowledge $E$. The Conditioning Prompt $P_{V}$ is used to generate \textit{GPT-4o} labels for SFT (the first stage) by providing the ground-truth $V^*$, it is also adopted to lead LLMs for preference pairs in the self-improvement DPO stage.

We list our prompt and introduce them in details as follows:

\begin{dialogbox}[The Veracity Generation Prompt $P_{CE}$ for the \textit{w/o} NEI Setting]
\textcolor{Red}{System:}\\
Task: Validate the following claim using the provided context. 

Your goal is to determine whether the claim can be supported by the context. Choose between "support" or "refute".

Instructions:
1. Analyze the claim step by step, verifying each crucial component in the claim as they appear.

2. Structure your reasoning on crucial components in the claim in detailed steps, from 1 to a maximum of 10. Make sure each step is the smallest possible logical unit necessary for validation.

3. Ensure that your reasoning correlates consistently with your conclusion. Use "\#\#" to format each step clearly, e.g., "\#\# Reasoning Step 1".

4. Finally, conclude with either "support" or "refute" enclosed in a pair of curly braces, noting the overall judgment regarding the claim.

~\\
\textcolor{Red}{User:}

\textcolor{RoyalBlue}{Context}: \"Life Goes On\" is a song recorded by American singer Fergie for her second studio album, \"Double Dutchess\" (2017). It was released as single on November 11, 2016, by Interscope and will.i.am Music Group. The song serves as the third single from Fergie's second studio album, following \"M.I.L.F. \$\". \"Life Goes On\" was written by Fergie, Tristan Prettyman, Keith Harris and Toby Gad.

\"M.I.L.F. \$\" (pronounced \"MILF money\") is a song recorded by American singer Fergie for her second studio album, \"Double Dutchess\" (2017). It was produced by Polow da Don and released as the second single from the record following \"L.A. Love (La La)\" on July 1, 2016 by Interscope and will.i.am Music Group. It debuted at number 34 on the US \"Billboard\" Hot 100 with 65,000 in first-week sales.
  
~\\

\textcolor{Gray}{Claim}: The song recorded by Fergie that was produced by Polow da Don and was followed by Life Goes On was M.I.L.F.\$.
\end{dialogbox}

We select a claim in the HOVER dataset to provide the example above for the \textit{w/o} NEI setting. The above-mentioned prompt is used in:
\begin{itemize}
    \item In training, providing claims and knowledge with CoT instructions (is used together with $P_V$).
    \item In testing, generating \textit{CoT-Verify} results for data under the \textit{w/o} NEI setting.
\end{itemize}

\begin{dialogbox}[The Veracity Generation Prompt $P_{CE}$ for the \textit{w} NEI Setting]
\textcolor{Red}{System:}\\
Task: Validate the following claim using the provided context. 

Your goal is to determine whether the claim can be supported with the context. Choose between "support", "refute", or "not enough information".

Instructions:

1. Analyze the claim step by step, verifying each crucial component in the claim as they appear.

2. Structure your reasoning on crucial components in the claim in detailed steps, from 1 to a maximum of 10. Make sure each step is the smallest possible logical unit necessary for validation.

3. Ensure that your reasoning correlates consistently with your conclusion. Use "\#\#" to format each step clearly, e.g., "\#\# Reasoning Step 1".

4. Finally, conclude with "support", "refute", or "not enough information" enclosed in a pair of curly braces, noting the overall judgment regarding the claim.

~\\
\textcolor{Red}{User:}

\textcolor{RoyalBlue}{Context}: In this study, we collected blood from COVID-19 patients who have recently become virus-free and therefore were discharged, and analyzed their SARS-CoV-2-specific antibody and T cell responses. We observed SARS-CoV-2-specific humoral and cellular immunity in the patients. Both were detected in newly discharged patients, suggesting both participate in immune-mediated protection to viral infection.
  
~\\

\textcolor{Gray}{Claim}: For most patients, COVID-19 begins and ends in their lungs, because like the flu, coronaviruses are respiratory diseases.
\end{dialogbox}

We select a claim in the Healthver dataset to provide the example above for the \textit{w} NEI setting. The above-mentioned prompt is used in:
\begin{itemize}
    \item In training, providing claims and knowledge with CoT instructions (is used together with $P_V$).
    \item In testing, generating \textit{CoT-Verify} results for data under the \textit{w} NEI setting.
\end{itemize}

\begin{dialogbox}[With the Ground-truth Veracity Prompt $P_{V}$ (the \textit{w/o} NEI Setting Case as an Example)]
\textcolor{Red}{System:}\\
Task: Validate the following claim using the provided context. 

Your goal is to determine whether the claim can be supported by the context. Choose between "support" or "refute".

Instructions:
1. Analyze the claim step by step, verifying each crucial component in the claim as they appear.

2. Structure your reasoning on crucial components in the claim in detailed steps, from 1 to a maximum of 10. Make sure each step is the smallest possible logical unit necessary for validation.

3. Ensure that your reasoning correlates consistently with your conclusion. Use "\#\#" to format each step clearly, e.g., "\#\# Reasoning Step 1".

4. Finally, conclude with either "support" or "refute" enclosed in a pair of curly braces, noting the overall judgment regarding the claim.

The ground truth is

---

Answer: {support}. You must generate results that match ground truth.

~\\
\textcolor{Red}{User:}

\textcolor{RoyalBlue}{Context}: \"Life Goes On\" is a song recorded by American singer Fergie for her second studio album, \"Double Dutchess\" (2017). It was released as single on November 11, 2016, by Interscope and will.i.am Music Group. The song serves as the third single from Fergie's second studio album, following \"M.I.L.F. \$\". \"Life Goes On\" was written by Fergie, Tristan Prettyman, Keith Harris and Toby Gad.

\"M.I.L.F. \$\" (pronounced \"MILF money\") is a song recorded by American singer Fergie for her second studio album, \"Double Dutchess\" (2017). It was produced by Polow da Don and released as the second single from the record following \"L.A. Love (La La)\" on July 1, 2016 by Interscope and will.i.am Music Group. It debuted at number 34 on the US \"Billboard\" Hot 100 with 65,000 in first-week sales.
  
~\\

\textcolor{Gray}{Claim}: The song recorded by Fergie that was produced by Polow da Don and was followed by Life Goes On was M.I.L.F.\$.
\end{dialogbox}

We select a claim in the HOVER dataset to provide the example above for $P_V$. This prompt is used in the two stages of training, providing the ground-truth $V^*$ (noted as $P_{V^*}$) in the first stage and guiding the conditional generation for each veracity option (noted as $P_{support}$, $P_{NEI}$, and $P_{refute}$) in the second stage.

\section{Training Dataset Examples}\label{example}
To help the comprehension of the Reasoning-CV fine-tuning process, in this section, we provide examples for the training data in two stages as follows:

\begin{dialogbox}[Example of SFT Training data in stage 1. (Output of \textbf{GPT-4o} with CoT prompt ($P_{CE}$) and conditioning prompt ($P_{V^*}$))]

\textcolor{Red}{"instruction":} Task: Validate the following claim using the provided context. 

Your goal is to determine whether the claim can be supported by the context. Choose between "support" or "refute".

Instructions:
1. Analyze the claim step by step, verifying each crucial component in the claim as they appear.

2. Structure your reasoning on crucial components in the claim in detailed steps, from 1 to a maximum of 10. Make sure each step is the smallest possible logical unit necessary for validation.

3. Ensure that your reasoning correlates consistently with your conclusion. Use "\#\#" to format each step clearly, e.g., "\#\# Reasoning Step 1".

4. Finally, conclude with either "support" or "refute" enclosed in a pair of curly braces, noting the overall judgment regarding the claim.

~\\

\textcolor{Red}{"input":} \textcolor{RoyalBlue}{Context}: Joseph Nassise (born 1968) is a New York Times and USA Today bestselling American urban fantasy writer and the author of more than forty novels. 

Joseph Nassise lives with his wife and four children in Arizona. Joseph Nassise is the author of the internationally bestselling Templar Chronicles series, the Jeremiah Hunt Chronicle, the Great Undead War series, as well as several books for Gold Eagle's Rogue Angel line. 

Joseph Nassise served as the president of the Horror Writers Association from 2002 to 2005 and a Trustee of the same from 2008 to 2010.

\textcolor{Gray}{Claim}: Joseph Nassise is a New York Times and USA Today bestselling American urban fantasy writer and the author residing in Arizona who wrote Templar Chronicles series, the Jeremiah Hunt Chronicl and the Great Undead War series and erved as the president of the Horror Writers Association from 2002 to 2005.

~\\

\textcolor{Green}{"output"}: "\#\# Reasoning Step 1: Verify if Joseph Nassise is a New York Times and USA Today bestselling American urban fantasy writer.

- The context states that Joseph Nassise is a New York Times and USA Today bestselling American urban fantasy writer.

\#\# Reasoning Step 2: Verify if Joseph Nassise resides in Arizona.

- The context confirms that Joseph Nassise lives with his wife and four children in Arizona.

\#\# Reasoning Step 3: Verify if Joseph Nassise wrote the Templar Chronicles series.

- The context mentions that Joseph Nassise is the author of the internationally bestselling Templar Chronicles series.

\#\# Reasoning Step 4: Verify if Joseph Nassise wrote the Jeremiah Hunt Chronicle.

- The context states that Joseph Nassise is the author of the Jeremiah Hunt Chronicle.

\#\# Reasoning Step 5: Verify if Joseph Nassise wrote the Great Undead War series.

- The context confirms that Joseph Nassise is the author of the Great Undead War series.

\#\# Reasoning Step 6: Verify if Joseph Nassise served as the president of the Horror Writers Association from 2002 to 2005.

- The context indicates that Nassise served as the president of the Horror Writers Association from 2002 to 2005.

\#\# Reasoning Step 7: Verify if there are any discrepancies or missing components in the claim.

- The claim accurately reflects the information provided in the context without any discrepancies or missing components.

Based on the above steps, the claim is fully supported by the context.

\{support\}"
\end{dialogbox}

\begin{dialogbox}[Example of DPO pairs in stage 2.]

\textcolor{Red}{"instruction":} Task: Validate the following claim using the provided context. 

Your goal is to determine whether the claim can be supported by the context. Choose between "support" or "refute".

Instructions:
1. Analyze the claim step by step, verifying each crucial component in the claim as they appear.

2. Structure your reasoning on crucial components in the claim in detailed steps, from 1 to a maximum of 10. Make sure each step is the smallest possible logical unit necessary for validation.

3. Ensure that your reasoning correlates consistently with your conclusion. Use "\#\#" to format each step clearly, e.g., "\#\# Reasoning Step 1".

4. Finally, conclude with either "support" or "refute" enclosed in a pair of curly braces, noting the overall judgment regarding the claim.

~\\

\textcolor{Red}{"input":} \textcolor{RoyalBlue}{Context}: The Boxer is a medium-sized, short-haired breed of dog, developed in Germany. The coat is smooth and tight-fitting; colours are fawn or brindled, with or without white markings, and white. Boxers are brachycephalic (they have broad, short skulls), have a square muzzle, mandibular prognathism (an underbite), very strong jaws, and a powerful bite ideal for hanging on to large prey. The Boxer was bred from the Old English Bulldog and the now extinct Bullenbeisser which became extinct by crossbreeding rather than by a decadence of the breed. The purpose of the crossbreeding was the wish to eliminate the excessive white color of the breed, and the necessity of producing thousands of dogs for one of the most popular breeds in the world. The Boxer is part of the Molosser group. This group is a category of solidly built, large dog breeds that all descend from the same common ancestor, the large shepherd dog known as a Molossus. The Boxer is a member of the Working Group.

Can de Palleiro or Pastor Galego also known as Galician Shepherd Dog or Galician Palleiro is a dog breed originating in Galicia (Spain).

The Bullenbeisser (also known as the German Bulldog) was a breed of dog known for its strength and agility. The breed was closely related to the Bärenbeisser (some believe that the two breeds were the same; the names mean \"bull-biter\" and \"bear-biter,\" respectively), and the Boxer. It was, in all its aspects, similar to the present Alano Español (Spanish Bulldog) and very alike to the Dogo Argentino, not only in aspect, but also in usage. There were two regional varieties, the Brabanter Bullenbeisser and the Danziger Bullenbeisser. The breed is now extinct.

\textcolor{Gray}{Claim}: The Can de Palleiro is a dog that originated in Wong, not the dog that the Old English Bulldog was bred with to make the Boxer (dog).

~\\

\textcolor{Green}{"chosen"}: "\#\# Reasoning Step 1: The claim states that the Can de Palleiro is a dog that originated in Wong. 

- The context specifies that the Can de Palleiro, also known as the Galician Shepherd Dog or Galician Palleiro, is a dog breed originating in Galicia, Spain, not Wong.

\#\# Reasoning Step 2: The claim states that the Can de Palleiro is not the dog that the Old English Bulldog was bred with to make the Boxer.

- The context indicates that the Boxer was bred from the Old English Bulldog and the now extinct Bullenbeisser, not the Can de Palleiro.

\#\# Reasoning Step 3: Verify the relationship between the Can de Palleiro and the Boxer.

- The context does not mention any relationship between the Can de Palleiro and the Boxer, further supporting that the Can de Palleiro was not involved in the breeding of the Boxer.

\#\# Reasoning Step 4: Confirm the accuracy of the claim based on the context.

- The context supports the claim that the Can de Palleiro is not the dog that the Old English Bulldog was bred with to make the Boxer, as it was the Bullenbeisser that was used.

\#\# Reasoning Step 5: Evaluate the overall claim.

- The claim is partially incorrect regarding the origin of the Can de Palleiro, but correct in stating that it was not involved in the breeding of the Boxer.

Based on the analysis, the claim is partially correct but contains an incorrect statement about the origin of the Can de Palleiro. Therefore, the claim is overall refuted. 

\{refute\}",

\textcolor{Green}{"rejected"}: "\#\# Reasoning Step 1: The claim states that the Can de Palleiro is a dog that originated in Wong. The context does not mention Wong as the origin of the Can de Palleiro. Instead, it states that the Can de Palleiro, also known as the Galician Shepherd Dog, originated in Galicia, Spain.

\#\# Reasoning Step 2: The claim states that the Can de Palleiro is not the dog that the Old English Bulldog was bred with to make the Boxer. The context specifies that the Boxer was bred from the Old English Bulldog and the now extinct Bullenbeisser, not the Can de Palleiro.

\#\# Reasoning Step 3: The context confirms that the Bullenbeisser was one of the breeds used to create the Boxer, and it was not the Can de Palleiro. Therefore, the claim that the Can de Palleiro was not used in the breeding of the Boxer is consistent with the context.

\#\# Reasoning Step 4: The context does not provide any information linking the Can de Palleiro to the breeding of the Boxer, further supporting the claim that it was not involved in the creation of the Boxer breed.

\#\# Reasoning Step 5: The claim is consistent with the context in stating that the Can de Palleiro is not related to the breeding of the Boxer, as the context clearly attributes the breeding of the Boxer to the Old English Bulldog and the Bullenbeisser.

\{support\}"
\end{dialogbox}

\section{Case Study}\label{case_study}

As the complement to Figure \ref{fig:error}, in this section, we show the cases of the LLM's drawbacks after the first stage of SFT fine-tuning.

\begin{dialogbox}[Case 1: Fault in judging the acceptance threshold.]\label{case_study1}

\textcolor{Gray}{Claim}: \textbf{Many} of the works of Dallas John Baker are gay-themed, though not all of them, and some of his short stories are set in Australia.

\textcolor{RoyalBlue}{Context}: Dallas John Baker (born Dallas John Angguish Baker on 19 February 1968) is an Australian writer and academic noted for poetry, short stories and travel writing whose work deals with themes of alienation, otherness and sexuality. \textbf{Some} have gay content, but not all of them. Local readers will be interested in the final section of stories, set in Australia with locales ranging from Byron Bay to the Darling Downs.

\textcolor{Red}{Ground-truth Veracity}: Support

\textcolor{Green}{Context}: \#\# Reasoning Step 1: The claim states that many of the works of Dallas John Baker are gay-themed. The context mentions that some of his works have gay content, but it does not specify that many of his works are gay-themed. Therefore, this part of the claim is not fully supported by the context.

...

\{refute\}

\end{dialogbox}

In this case, the LLM model determines a claim that can be validated by context knowledge as \textit{`refute`}. Based on the output content, we can find that the wrong output is because the LLM believes that the paraphrase from ``some`` to ``many`` can be a fundamental fault, which does not match the human judging threshold. Therefore, we design Stage 2 of Reasoning-CV to align the LLM judgment with a reasonable threshold.

\begin{dialogbox}[Case 2: Inconsistent reasoning paths.]\label{case_study2}

\textcolor{Gray}{Claim}: The suburb of Lousiville is the headquarters of the business founded by John H Schnatter. It is a rule-class city of 26,595 and it contains the neighborhood Plainview.

\textcolor{RoyalBlue}{Context}: Plainview is a neighborhood in the city of Jeffersontown, Kentucky and is located on the former site of the Plainview Dairy Farm. It is a Planned Unit Development which includes housing, workplaces, and shopping within a single development.

John H. Schnatter (born November 22, 1961) is an American entrepreneur and the founder, CEO, and spokesman of Papa John's International, Inc.

Jeffersontown is a home rule-class city in Jefferson County, Kentucky, United States. The population was 26,595 at the 2010 U.S. census.

Papa John's Pizza is an American restaurant franchise company. It runs the third largest take-out and pizza delivery restaurant chain in the United States, with headquarters in Jeffersontown, Kentucky, a suburb of Louisville.

\textcolor{Red}{Ground-truth Veracity}: Support

\textcolor{Green}{Context}: 

\#\# Reasoning Step 2: From the Claim, Jeffersontown is a suburb of Louisville, Kentucky.

...

\#\# Reasoning Step 5: The claim correctly states that the headquarters of the business founded by John H. Schnatter is in Jeffersontown, which contains the neighborhood Plainview. However, it incorrectly identifies Louisville as the suburb.

\{refute\}

\end{dialogbox}

In this case, the LLM model also determines a claim that can be validated by context knowledge as \textit{`refute`}. Based on the output content, we can find that the wrong output is because the LLM performs a hallucination and suddenly changes its previous correct assertion ``Jeffersontown is a suburb of Louisville`` to ``it incorrectly identifies Louisville as the suburb``. Therefore, we design stage 2 of Reasoning-CV to improve the consistency of the \textit{CoT-Verify} generation.

\newpage

\twocolumn

\section{Details of Reasoning-CV}

In this section, we provide necessary details of the proposed Reasoning-CV, including the fine-tuning procedure for the \textit{w} dataset Healthver.

\subsection{Fine-Tuning Details for Claims in Healthver} \label{nei_SI}

In the training dataset of Reasoning-CV, FEVEROUS and HOVER support only the \textit{w/o} NEI setting (i.e., have \textit{`support`} and \textit{`refute`} options in their labels), and Healthver supports the \textit{w} NEI setting (i.e., has \textit{`support`}, \textit{`NEI`}, and \textit{`refute`} options in its labels).

So, in the second stage of Reasoning-CV, we enumerate two options for FEVEROUS and HOVER (as shown in the main text), and enumerate three options for Healthver. The detailed process of building $\mathcal{D}_{SI_{i-1}}$ for Healthver is as follows:
\begin{equation}
\begin{aligned}
&R_{support} = LLM_{SI_{i-1}}(P_{CE}, P_{support}),\\&R_{NEI} = LLM_{SI_{i-1}}(P_{CE}, P_{NEI}), \\&R_{refute} = LLM_{SI_{i-1}}(P_{CE}, P_{refute}).
\end{aligned}\label{Stage2-nei}
\end{equation}
\begin{equation}
\begin{aligned}
\mathcal{D}_{SI_{i-1}} = \Big\{&[V^* \in R_{support}](R_{refute}\prec R_{support}), \\&[V^* \in R_{support}](R_{NEI}\prec R_{support}), \\&[V^* \in R_{NEI}](R_{refute}\prec R_{NEI}), \\&[V^* \in R_{NEI}](R_{support}\prec R_{NEI}), \\&[V^* \in R_{refute}](R_{support}\prec R_{refute}), \\&[V^* \in R_{support}](R_{NEI}\prec R_{refute})\Big\},
\end{aligned}
\end{equation}

% \subsection{Details for Open Book Experiments}

\begin{figure*}[htbp]
\centering
\includegraphics[width=0.9\linewidth]{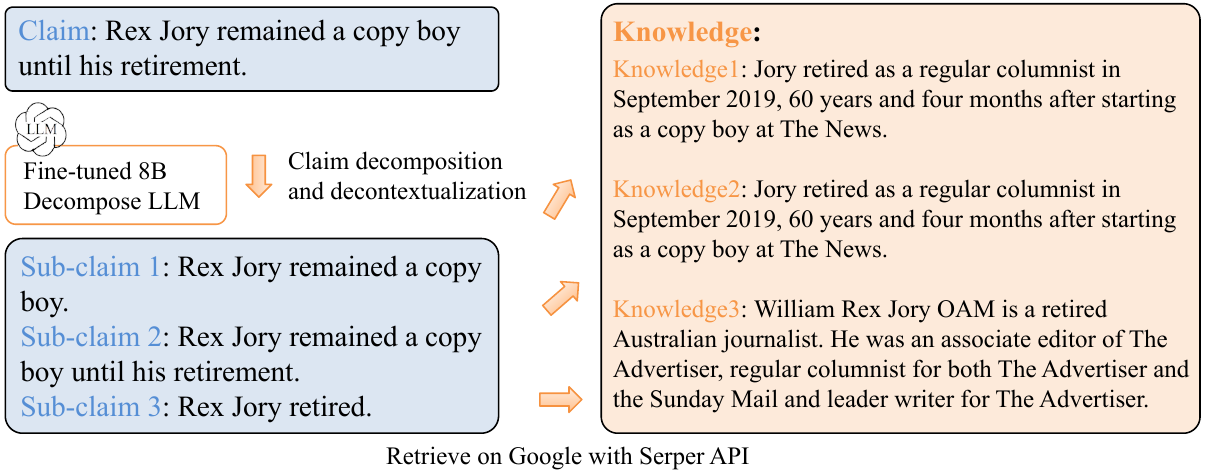}
\caption{The evidence retrieval process in the open book setting of Reasoning-CV. We fine-tune another 8B LlaMA model for the claim decomposition and decontextualization tasks in \citet{min2023factscore}. We retrieve the top paragraph from Google by the Serper API, taking each sub-claim as a query. }
\label{fig:open_book}
\end{figure*}

\section{More Experiments}

\subsection{Reasoning-CV Performance In the Open Book Setting} \label{open_book_result}

In the Open Book setting of adopted datasets, the claim verification method is required to retrieve knowledge $E$ from the Internet \cite{wang2023explainable}. To achieve this, as shown in Figure \ref{fig:open_book}, we follow the idea from \citet{zhao2024pacar}, break the claim into sub-claims, and then retrieve the most related paragraph from Google with the Serper API, taking each sub-claim as the query. We use the prompts from \citet{min2023factscore} for the claim decomposition and decontextualization. To build an 8B open book claim verification system, we fine-tune another \textit{Meta-LlaMA-3-8B-Instruct} LLM by SFT based on the \textit{GPT-4o} output of claim decomposition and decontextualization for independent sub-claims. As shown in Table \ref{result_nei_open}, where we use the same collected knowledge for all baselines except \citet{zhao2024pacar}. When using the knowledge $E$ obtained from the open book setting, Reasoning-CV still outperforms \textit{GPT-4o} in the average performance, demonstrating the generalization ability of the obtained model to various evidence settings.

\begin{table*}[htbp]
\centering
\renewcommand\arraystretch{1}
\setlength{\tabcolsep}{1.5mm}
\resizebox{1\textwidth}{!}{
\begin{tabular}{lcccccccc}
\bottomrule[0.5mm]
\multicolumn{9}{c}{Open Book} \\ \hline
\multicolumn{1}{c|}{Dataset}   & \multicolumn{1}{c|}{Param Size} & \multicolumn{1}{c|}{FEVEROUS}& HOVER (2-hop)   & HOVER (3-hop)   & \multicolumn{1}{c|}{HOVER (4-hop)}   & \multicolumn{2}{c|}{Healthver}& \multicolumn{1}{l}{} \\ \cline{1-8}
\multicolumn{1}{c|}{Method}& \multicolumn{1}{c|}{Settings}   & \multicolumn{1}{c|}{\textit{w/o NEI}} & \textit{w/o NEI}& \textit{w/o NEI}& \multicolumn{1}{c|}{\textit{w/o NEI}}& \textit{w/o NEI}& \multicolumn{1}{c|}{\textit{w NEI}}  & \textit{Avg.}  \\ \hline
\multicolumn{1}{l|}{GPT-4o}& \multicolumn{1}{c|}{N/A}   & \multicolumn{1}{c|}{\underline{74.27}} & 68.37 & 57.06 & \multicolumn{1}{c|}{52.42} & \underline{55.30} & \multicolumn{1}{c|}{\underline{43.27}} & \underline{58.45}\\
\multicolumn{1}{l|}{Minicheck-7B}   & \multicolumn{1}{c|}{7B}& \multicolumn{1}{c|}{74.03}  & 65.90 & 46.75 & \multicolumn{1}{c|}{40.90} & 52.47 & \multicolumn{1}{c|}{-}& 56.01 \\ \hline
\multicolumn{9}{c}{Estinblished \textit{Decompose-Then-Verify} Methods}   \\ \hline
\multicolumn{1}{l|}{PACAR*}& \multicolumn{1}{c|}{N/A}   & \multicolumn{1}{c|}{72.61}  & \underline{\textbf{73.13}} & \underline{\textbf{64.07}} & \multicolumn{1}{c|}{\underline{\textbf{63.82}}} & -& \multicolumn{1}{c|}{-}& -\\
\multicolumn{1}{l|}{FactScore} & \multicolumn{1}{c|}{N/A}   & \multicolumn{1}{c|}{69.29}  & 61.10 & 46.11 & \multicolumn{1}{c|}{43.50} & 53.86 & \multicolumn{1}{c|}{41.40} & 54.77 \\
\multicolumn{1}{l|}{Decompose + Minicheck-7B} & \multicolumn{1}{c|}{7B}& \multicolumn{1}{c|}{69.87}  & 61.14 & 41.71 & \multicolumn{1}{c|}{36.06} & 51.98 & \multicolumn{1}{c|}{-}& 52.15 \\ \hline
\multicolumn{9}{c}{Reasoning-CV for High-Qualty \textit{CoT-Verify} } \\ \hline
\multicolumn{1}{l|}{Meta-LLaMA-3-8B-Instruct} & \multicolumn{1}{c|}{8B}& \multicolumn{1}{c|}{78.80}  & 65.75 & 60.06 & \multicolumn{1}{c|}{58.44} & 48.93 & \multicolumn{1}{c|}{32.36} & 57.39 \\
\multicolumn{1}{l|}{+ CoT Prompt}   & \multicolumn{1}{c|}{8B}& \multicolumn{1}{c|}{76.53}  & 63.59 & 53.98 & \multicolumn{1}{c|}{54.92} & 54.38 & \multicolumn{1}{c|}{36.82} & 56.70 \\
\multicolumn{1}{l|}{+ Reasoning-CV-$LLM_{SFT}$}  & \multicolumn{1}{c|}{8B}& \multicolumn{1}{c|}{79.86}  & 72.29 & 59.16 & \multicolumn{1}{c|}{51.24} & \textbf{55.59} & \multicolumn{1}{c|}{41.88} & 60.00 \\
\multicolumn{1}{l|}{+ Reasoning-CV-$LLM_{SI_1}$}& \multicolumn{1}{c|}{8B}& \multicolumn{1}{c|}{\textbf{80.32}}  & 72.47 & 59.64 & \multicolumn{1}{c|}{49.20} & 54.20 & \multicolumn{1}{c|}{\textbf{43.35}} & \textbf{59.86}  \\
\multicolumn{1}{l|}{+ Reasoning-CV-$LLM_{SI_2}$}& \multicolumn{1}{c|}{8B}& \multicolumn{1}{c|}{80.22}   & 71.98 & 61.44 & \multicolumn{1}{c|}{49.86} & 54.05 & \multicolumn{1}{c|}{41.60} &\textbf{59.86} \\ \toprule[0.5mm]
\end{tabular}}
\caption{Reasoning-CV fine-tuned LLMs in in-domain test sets with open book considering both \textit{w NEI} and \textit{w/o NEI} settings. We \textbf{bold} the best-performing LLM on each test set and \underline{underlines} the best-performing baseline. Results with * are collected from \citet{zhao2024pacar}, with possibly different knowledge for each claim.}
\label{result_nei_open}
\end{table*}

\subsection{Comparison to Claim Verification Methods with Adaptive Retrieval}

As mentioned in related work, ProgramFC \cite{pan2023fact} is a typical claim verification method with adaptive retrieval. In this part, we compare the proposed Reasoning-CV fine-tuned LLMs with ProgramFC, as shown in Table \ref{program}, Reasoning-CV leads in 7 out of 8 test sets, demonstrating the power of the proposed Reasoning-CV and its adopted \textit{CoT-Verify} paradigm.

\begin{table*}[htbp]
\centering
\setlength{\tabcolsep}{2mm}
\renewcommand\arraystretch{1}
\resizebox{\textwidth}{!}{
\begin{tabular}{l|cccc|cccc|c}
\bottomrule[0.5mm]
& \multicolumn{4}{c|}{Gold Evidence}  & \multicolumn{4}{c|}{Open Book} & \multicolumn{1}{l}{}  \\ \hline
\multicolumn{1}{c|}{\multirow{2}{*}{Method}} & \multirow{2}{*}{FEVEROUS} & \multicolumn{3}{c|}{HOVER} & \multirow{2}{*}{FEVEROUS} & \multicolumn{3}{c|}{HOVER} & \multirow{2}{*}{\textit{Avg.}} \\ \cline{3-5} \cline{7-9}
\multicolumn{1}{c|}{}&   & 2-hop& 3-hop& 4-hop&   & 2-hop& 3-hop& 4-hop& \\ \hline
GPT-4o   & 91.64& 79.48& 75.83& 73.52& 74.27& 68.37& 57.06& 52.42& 71.57\\
ProgramFC*& 91.77& 74.10& 66.13& 65.69& 67.80& 69.36& 60.63& \textbf{59.16} & 69.33\\ \hline
Meta-LLaMA-3-8B-Instruct& 90.34& 74.24& 67.21& 66.39& 78.80& 65.75& 60.06& 58.44& 70.15\\
+ Reasoning-CV-$LLM_{SFT}$   & 93.72& 81.26& 78.70& 75.65& 79.86& \textbf{72.29}& 59.16& 51.24& 73.99\\
+ Reasoning-CV-$LLM_{SI_2}$ & \textbf{95.50}& \textbf{85.97} & \textbf{83.93} & \textbf{83.02} & \textbf{80.22}& 71.98 & \textbf{61.44} & 49.86& \textbf{76.49}   \\ \toprule[0.5mm]
\end{tabular}}
\caption{Reasoning-CV fine-tuned LLMs in in-domain test sets with both gold evidence and open book. We bold the best result for each test set. We use the report results for ProgramFC. }
\label{program}
\end{table*}

\begin{table*}[htbp]
\centering
\setlength{\tabcolsep}{1mm}
\renewcommand\arraystretch{1.1}
\resizebox{\textwidth}{!}{
\begin{tabular}{c|cc|cc|cc|c}
\bottomrule[0.5mm]
 & \multicolumn{2}{c|}{Stage1} & \multicolumn{2}{c|}{Stage2-Round1} & \multicolumn{2}{c|}{Stage2-Round2} &   \\ \cline{1-7}
Base LLM & \textit{GPT-4o} labeling & LlaMA SFT & $LLM_{SFT}$ labeling & $LLM_{SFT}$ DPO & $LLM_{SI_1}$ labeling & $LLM_{SI_1}$ DPO & Total \\ \hline
LLaMA-3.2-3B-Instruct& 12h & 6h& 6h & 3.2h  & 7h & 3.2h  & 37.4h \\
Meta-LLaMA-3-8B-Instruct & 12h & 8.5h  & 3h & 7.5h  & 3.2h   & 9h& 43.2h \\ \toprule[0.5mm]
\end{tabular}}
\caption{Reasoning-CV fine-tuned LLMs in in-domain test sets with both gold evidence and open book. We bold the best result for each test set. We use the report results for ProgramFC. }
\label{time}
\end{table*}

\begin{figure}[H]
    \centering
    \includegraphics[width=1\linewidth]{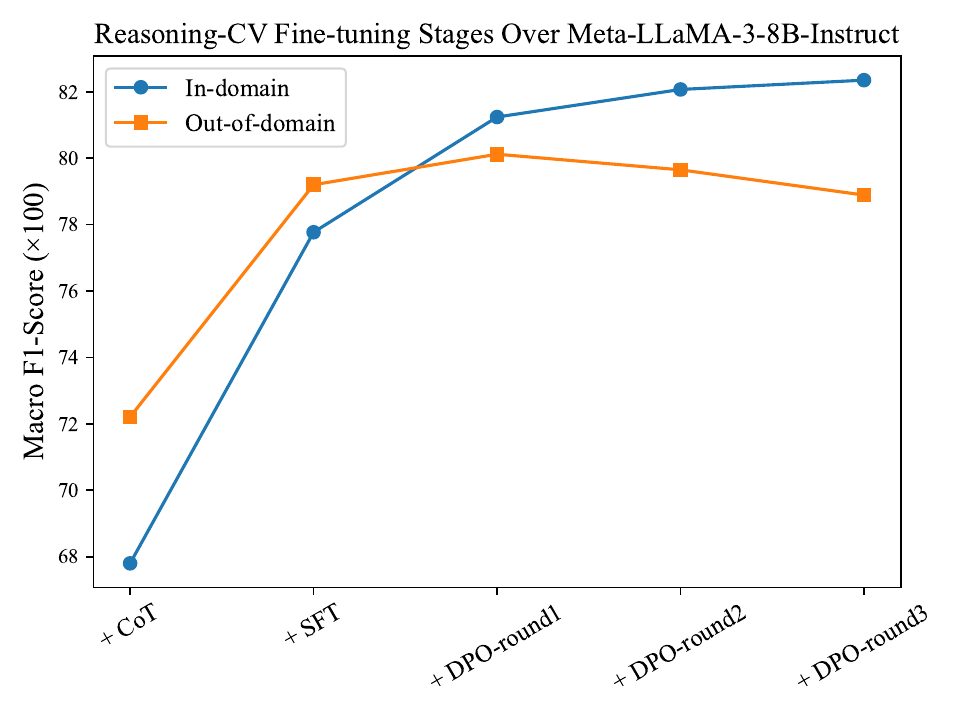}
    \caption{The performance curves for in-domain and out-of-domain datasets as the Reasoning CV fine-tuning progresses. }
    \label{fig:trend}
\end{figure}

\subsection{Performance over the Number of Rounds for Self-improvement DPO}
Table \ref{result_nei} and Table \ref{OOD} preliminarily demonstrate the performance changes of Reasoning-CV fine-tuned LLMs as the number of self-improvement DPO rounds increases. In this section, we use the average performance over dataset and settings (i.e., ``Avg.`` in Table \ref{result_nei} and Table \ref{OOD}) to evaluate the in-domain and out-of-domain performances, respectively, and draw a performance curve over Reasoning-CV stages and DPO rounds in Figure \ref{fig:trend}. As the number of DPO rounds increases from 0 (i.e., + SFT) to 3, Reasoning CV fine-tuned LLMs will perform significantly better in the in-domain. However, more than one round of Self-improvement DPO may impair LLM's effectiveness on out-of-domain claims, and \textit{Meta-LLaMA-3-8B-Instruct} + Reasoning-CV-$LLM_{SI_1}$ can be the best claim verification LLM for the out-of-domain.

\subsection{Time Consumption for Stages}

To evaluate the efficiency of Reasoning-CV finetuning, we list the time consumption for Reasoning-CV on a single H100-96B GPU. As shown in Table \ref{time}, the entire Reasoning-CV fine-tuning process can be completed in a relatively short amount of time (which will take less than two days).

\section{Dataset \& Baseline \& Licenses}

This section will list all the datasets and baseline methods used in this article, introducing their features and implementation details, and providing their sources and licenses.

\begin{table*}[htbp]
\centering
\renewcommand\arraystretch{1}
\setlength{\tabcolsep}{4mm}
\resizebox{\textwidth}{!}{
\begin{tabular}{l|ccccc}
\bottomrule[0.5mm]
 & \multicolumn{5}{c}{In-Domain} \\ \hline
Dataset& FEVEROUS & HOVER(2-hop)  & HOVER(3-hop) & HOVER(4-hop) & Healthver\\ \hline
\# of Eval & 2962& 1126& 1835& 1039& 1823\\ \hline
 & \multicolumn{5}{c}{Out-of-Domain}  \\ \hline
Dataset& LLM-AggreFact & SciFact-train & SciFact-dev  & VitaminC-dev & VitaminC-test \\ \hline
\# of Eval & 29320& 809 & 300& 63054   & 55197\\ \toprule[0.5mm]
\end{tabular}}
\caption{The number of claims in each dataset.}\label{claimnumber}
\end{table*}
\subsection{Dataset} \label{dataset_info}

This paper involves three datasets with complex claims (i.e., FEVEROUS, HOVER, and Healthver) and three datasets with relatively simple claims (i.e., LLM-AggreFact, SciFact, and VitaminC). 

Healthver, SciFact, and VitaminC support both the \textit{w} NEI setting and \textit{w/o} NEI setting (taking the label \textit{`NEI`} as a sub-case of \textit{`refute`}). 

\textbf{FEVEROUS}: FEVEROUS \cite{aly2021feverous} is a famous dataset for knowledge-assisted claim verification over unstructured and structured data, collecting evidence from sentences or cells from tables in Wikipedia. We adopt the setup from the previous methods \citet{pan2023fact, zhao2024pacar}, only considering claims that require sentence evidence.

\textbf{HOVER}: HOVER \cite{jiang2020hover} includes claims that can only be solved with multi-hop reasoning. It is divided into subsets based on the number of reasoning “hops” needed for claim verification (e.g., HOVER(2-hop) in Table \ref{result_nei} for claims 2-hop reasoning).

\textbf{Healthver}: Healthver \cite{Sarrouti2021Healthver} collects claims from real-world scenarios and knowledge from scientific articles. Unlike most claim verification datasets, where contradicted claims are usually just the negation of the supported ones, in Healthver, contradicted claims are themselves extracted from real-world claims, so the claims in this dataset are more challenging compared to other datasets \cite{jafari2024robust}. 

\textbf{LLM-AggreFact}: LLM-AggreFact dataset is a benchmark proposed in \citet{tang2024minicheck}, which is an aggregation of 10 existing datasets with relatively simple claims, including AggreFact \cite{tang2022understanding}, TofuEval \cite{tang2024tofueval}, ClaimVerify \cite{liu2023evaluating}, LGQA \cite{chen2023understanding}, ExpertQA \cite{malaviya2023expertqa}, Reveal \cite{jacovi2024chain}, FactCheck-GPT \cite{wang2023factcheck}, and WICE \cite{kamoi2023wice}. For this dataset, decomposition of sentences into atomic facts is not necessary to achieve this high performance \cite{tang2024minicheck}. \textbf{Importantly}, this dataset considers the scenario of detecting LLM output hallucination. Evaluating claim verification methods on this extensive dataset can effectively demonstrate their generalization ability.

\textbf{SciFact}: SciFact \cite{wadden2020fact} focuses on claims and knowledge in the science domain. We use the training set and the development set for evaluation because these sets provide gold evidence.

\textbf{VitaminC}: Evidence sources often change over time as more information is gathered and revised. To adapt to this change, models must be sensitive to subtle differences in supporting evidence. VitaminC \cite{schuster2021get} is proposed to evaluate the ability of claim verification methods in this situation. We consider both its development set and testing set for evaluation.

The number of claims in the test set of each dataset is listed in Table \ref{claimnumber}.

\begin{table*}[htbp]
\centering
\setlength{\tabcolsep}{1mm}
\renewcommand\arraystretch{1}
\resizebox{\textwidth}{!}{
\begin{tabular}{llll}
\bottomrule[0.5mm]
Resources & Type& License& URL\\ \midrule[0.3mm]
Minicheck & Code& Apache-2.0 license &  \url{https://github.com/Liyan06/MiniCheck} \\  
Minicheck-1B & LLM& MIT License &  \url{https://huggingface.co/lytang/MiniCheck-Flan-T5-Large} \\  
Minicheck-7B & LLM& CC BY-NC 4.0 &  \url{https://huggingface.co/bespokelabs/Bespoke-MiniCheck-7B} \\  \midrule[0.3mm]
FEVEROUS & Dataset& MIT License&\url{https://github.com/teacherpeterpan/ProgramFC}\\ 
HOVER & Dataset& MIT License&\url{https://github.com/teacherpeterpan/ProgramFC}\\ 
Healthver & Dataset& Available online&\url{https://github.com/sarrouti/Healthver} \\ 
LLM-AggreFact & Benchmark & CC BY-NC 4.0&\url{https://huggingface.co/datasets/lytang/LLM-AggreFact} \\ 
SciFact & Dataset& CC BY-NC 4.0 \& ODC-By 1.0   &\url{https://github.com/allenai/scifact?tab=readme-ov-file} \\ 
VitaminC & Dataset& CC BY-SA 3.0&\url{https://huggingface.co/datasets/tals/vitaminc} \\ \toprule[0.5mm]
\end{tabular}}
\caption{A summary of licenses.}
\label{1}
\end{table*}

\subsection{Baseline}\label{baseline_info}

This paper includes advanced black-box LLMs and established claim verification methods as baselines, including \textbf{Minicheck} \cite{tang2024minicheck}, \textit{\textbf{GPT-4o}}, \textbf{\textit{GPT-4o}+CoT}, and \textbf{o1-preview}, \textbf{FactScore} \cite{min2023factscore}, \textbf{PACAR} \cite{zhao2024pacar}, and \textbf{Decompose + Minicheck \cite{tang2024minicheck}}. We also include ProgramFC \cite{pan2023fact} in the Appendix.

We use the OpenAI API for \textit{\textbf{GPT-4o}}, \textbf{\textit{GPT-4o}+CoT}, and \textbf{o1-preview}. 

We reimplement the decomposition and decontextualization process in \textbf{FactScore} with the provided prompts \cite{min2023factscore} and validate the veracity of sub-claims by prompting LLMs with knowledge. Running FactScore usually requires tens of LLM calls for each complex claim, so we can only use \textit{GPT-4o-mini} to implement it. For the \textit{w/o} NEI setting, we judge the veracity of claims with Eq. \ref{folk}. For the \textit{w} NEI setting, with a similar spirit, we judge the claim as \textit{`NEI`} if and only if there are \textit{`NEI`} sub-claims but no \textit{`refute`} sub-claims.

For \textbf{Minicheck} \cite{tang2024minicheck}, we use their model for verification results.  \textbf{Minicheck} \cite{tang2024minicheck} suggests users to break complex claims first, so to break complex claims into easier ones, in Table \ref{result_nei}, we include \textbf{Decompose + Minicheck}. \textbf{Decompose + Minicheck} utilizes \textit{GPT-4o-mini} and prompts from FactScore \cite{min2023factscore} for claim decomposition and decontextualization. We observe consistent conclusions with \citet{hu2024decomposition} and \citet{tang2024minicheck} that breaking claims does not improve the effectiveness for claim verification, which reinforces the correctness of the adopted \textit{CoT-Verify} paradigm in Reasoning-CV. It is worth noting that on the dataset with complex claims (shown in Table \ref{result_nei}), we can observe the conclusion shown in \citet{tang2024minicheck}, a separate decomposition step will not lead to better claim verification results.

For \textbf{PACAR} \cite{zhao2024pacar}, it reaches the current state-of-the-art claim verification performances with the \textit{Decompose-Then-Verify} paradigm, but we cannot get their prompts and implementations. So, we report their results in articles \cite{zhao2024pacar}. 

\textbf{ProgramFC} \cite{pan2023fact} is a typical claim verification method for adaptive retrieval and solving ideas. We also use the report results in the Appendix for ProgramFC.

\subsection{License}
The licenses and URL of baselines are listed in Table \ref{1}.

\end{document}